\documentclass[letterpaper]{article} 
\usepackage{aaai2027} 
\usepackage{times}  
\usepackage{helvet}  
\usepackage{courier}  
\usepackage[hyphens]{url}  
\usepackage{graphicx} 
\usepackage{natbib}  
\usepackage{caption} 
\usepackage{amsfonts} 
\usepackage{amsmath}  
\usepackage{graphicx}
\usepackage{subcaption}
\usepackage{algorithm}
\usepackage{algorithmic}
\usepackage{booktabs}

\title{PCGRL for Complex Platformers: Synthesizing Playable and Diverse Levels via Inflated Envelopes and Forward Curricula}

\author{
    Di Wu\thanks{Currently with The University of Hong Kong. Correspondence to: wudi3686343@connect.hku.hk}
}
\affiliations{
    Harbin Institute of Technology
}
\nocopyright
\begin{document}

\maketitle

\begin{abstract}
Procedural Content Generation via Reinforcement Learning (PCGRL) has emerged as a promising paradigm for automated game development. However, existing PCGRL frameworks still suffer from severe exploration bottlenecks and scaling limitations when faced with long-sequence generation tasks governed by strict rigid-body rules and non-linear kinematic boundaries (e.g., platformer constraints). Consequently, traditional approaches frequently fail to maintain reliable level solvability on large-scale canvases. To address these challenges, we present a physics-aware, curriculum-guided PCGRL framework. First, we construct an "Inflated Volume Envelope" mechanism, which provides a physics-constrained representation that incorporates continuous kinematic feasibility into the discrete state space of the MDP. Second, to alleviate the sparse reward challenge induced by extremely long decision sequences, we introduce intermediate state-transition feedback and a multi-stage forward extension curriculum learning framework, enabling informative value estimation and stable credit assignment over long-horizon trajectories. Finally, to resolve the issue where deterministic policies after convergence generate highly monotonous map paths with a lack of diversity, we activate high-entropy stochastic inference during the synthesis phase, regularized by a post-hoc parallel rejection filtering mechanism. Experimental results demonstrate that our framework significantly enhances structural diversity while maintaining high physical playability across multi-scale environments. Comparative evaluations across multi-scale canvases demonstrate that our framework maintains superior convergence stability and dimensional scalability, reliably synthesizing solvable levels where traditional PCGRL baselines degrade.
\end{abstract}

\section{Introduction}
Modern game development increasingly explores Procedural Content Generation (PCG) to create large-scale and diverse player experiences. While traditional rule-based and search-based methods can provide strong playability constraints, they often require extensive domain-specific heuristics and manual engineering. In recent years, framing content generation as a sequential decision-making problem via Reinforcement Learning (PCGRL) has provided a new perspective for automated content generation, allowing autonomous agents to learn complex spatial distributions and layout topologies through environmental interactions.

However, generating structurally intricate, large-scale levels—particularly those governed by complex physical constraints and nonlinear kinematic boundaries (e.g., precise jumping mechanics in platformer games)—remains a formidable challenge. In these high-precision domains, valid playable trajectories occupy only a small fraction of the massive discrete design space. Consequently, conventional PCGRL agents suffer from inefficient exploration and unstable optimization when scaling to long-horizon generation, leading to degraded level solvability in large environments.

Furthermore, a critical trade-off exists between playability and diversity in current PCGRL frameworks. Even when agents successfully converge in highly constrained environments, they tend to collapse into deterministic, "safe" behavioral policies. This policy stagnation leads to the generation of monotonous and repetitive level paths, limiting the ability of PCG systems to produce diverse and engaging content. Striking a balance between strict kinematic feasibility and gameplay creativity remains an open question.

To address these challenges, we propose a physics-aware PCGRL framework that explicitly bridges continuous kinematic constraints with discrete level editing. Rather than forcing a single agent to implicitly learn both physics rules and aesthetic diversity under sparse rewards, our framework systematically decomposes the generation pipeline.

Specifically, we introduce a physics-aware state representation based on the Inflated Volume Envelope, a curriculum-guided optimization strategy with intermediate feedback, and a diversity-preserving stochastic synthesis mechanism.

The main contributions of this work are summarized as follows:

Physics-Aware State Representation: We propose the Inflated Volume Envelope mechanism, bridging the gap between continuous kinematic constraints and discrete PCGRL state spaces.

Long-Horizon Curriculum Engine: We design a multi-stage forward extension curriculum with intermediate feedback, effectively alleviating exploration difficulties and scalability limitations in large-canvas generation.

Diversity-Preserving Synthesis: We introduce a decoupled generation scheme utilizing high-entropy stochastic inference paired with parallel rejection filtering, reducing the loss of diversity caused by deterministic policy convergence.

Empirical Validation: Comprehensive evaluations across multi-scale canvases demonstrate that our framework maintains stable convergence and robust dimensional scalability, while stochastic synthesis preserves structural diversity without sacrificing physical playability.
\section{Related Work}

\subsection{Procedural Content Generation for Games}

Procedural Content Generation (PCG) aims to automatically create game assets, levels, and environments, reducing manual design effort while enabling scalable content creation \citep{togelius2011search, shaker2016procedural}. Early approaches based on handcrafted rules, search-based optimization, and evolutionary computation provide strong controllability but rely heavily on domain-specific heuristics, limiting scalability in complex environments with dynamic physical interactions.

To improve structural reliability, constraint-based and movement-aware PCG explicitly incorporate gameplay and kinematic requirements during synthesis. In platformer generation, systems must account for jump arcs and long-range reachability dependencies. Early statistical methods struggled under strict jump constraints, achieving low playability rates in \textit{Kid Icarus} \citep{snodgrass2016learning}; subsequent work explicitly incorporated player movement models into Markov-based level generation to improve physical traversability \citep{snodgrass2017player}. However, such discrete transition models still rely on handcrafted movement abstractions and lack end-to-end policy adaptability across arbitrary canvas scales.

\subsection{Learning-based PCG and PCGRL}

Learning-based Procedural Content Generation (PCGML) leverages deep generative models, such as VAEs, GANs, and large language models, to learn content distributions from existing level corpora \citep{snodgrass2016learning, summerville2016super, torrado2019deep, awiszus2020toad, sarkar2021dungeon, todd2023level, gallotta2024large}. To promote structural variety while preserving playability, Quality-Diversity (QD) frameworks further combine latent generative models with behavioral illumination to discover diverse sets of traversable levels \citep{fontaine2021illuminating, bhatt2022deep}. Despite their expressive capacity, data-driven methods depend heavily on pre-existing level datasets, and unconstrained generation may produce content that violates continuous physical reachability constraints.

To overcome data dependence, Procedural Content Generation via Reinforcement Learning (PCGRL) reformulates content creation as a Markov Decision Process (MDP), enabling agents to learn editing policies through environmental feedback \citep{khalifa2020pcgrl}. Subsequent extensions introduced controllable conditioning parameters \citep{earle2021learning}, scaled PCGRL to three-dimensional environments \citep{jiang2022learning}, and explored multi-agent decentralized editing formulations to improve map scalability \citep{earle2025video}. While PCGRL and sequence-based generators have demonstrated success in dungeon and platformer environments \citep{khalifa2020pcgrl, sudhakaran2023mariogpt}, standard RL-based generators typically rely on purely discrete tile observations, requiring a single policy to implicitly learn continuous physics rules and structural diversity from sparse feedback. Unlike prior methods that rely on point-wise discrete path masks or unconstrained end-to-end learning, our framework embeds a continuous swept-volume envelope directly into the MDP state and pairs it with a forward curriculum to stabilize large-canvas optimization.

\subsection{Exploration and Curriculum Learning in Reinforcement Learning}

Exploration remains challenging in reinforcement learning under sparse rewards and long-horizon decision processes~\cite{sutton2018reinforcement}. Curriculum learning improves optimization by gradually increasing task difficulty and introducing intermediate objectives~\cite{bengio2009curriculum, florensa2017reverse}. However, existing curriculum strategies rarely consider physical constraints in procedural content generation. Our work addresses this gap through a forward curriculum with intermediate state-transition feedback.

\begin{figure}[t] 
  \centering
  \includegraphics[width=\linewidth]{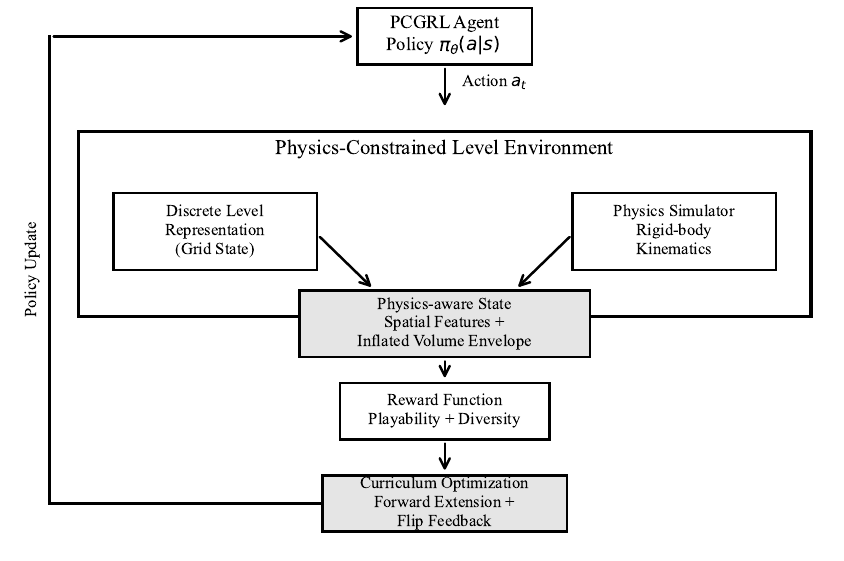} 
  \caption{    Overview of the proposed physics-aware PCGRL framework.
    The framework integrates discrete level editing with continuous
    kinematic verification through a physics-aware state representation,
    followed by reward-guided curriculum optimization.}
  \label{fig:framework}
\end{figure}

\section{Method}

\subsection{Problem Formulation and Overview}

We formulate physics-constrained level generation as a sequential decision-making problem under the PCGRL framework. Given an initially empty or partially initialized canvas, an agent iteratively edits the level representation through a sequence of actions, aiming to generate content that satisfies both physical feasibility and structural design objectives.

Specifically, the generation process is modeled as a Markov Decision Process (MDP):

\begin{equation}
\mathcal{M} = \langle \mathcal{S}, \mathcal{A}, \mathcal{R}, \mathcal{T}, \gamma \rangle
\end{equation}

where $\mathcal{S}$ denotes the state space describing the current level configuration, $\mathcal{A}$ represents the editing action space, $\mathcal{R}$ is the reward function evaluating generated structures, $\mathcal{T}$ denotes the environment transition dynamics, and $\gamma$ is the discount factor.

Unlike conventional PCGRL settings that treat level generation as purely discrete tile manipulation, our framework explicitly considers the interaction between discrete level structures and continuous physical dynamics. The key challenge is that a visually valid layout does not necessarily correspond to a physically playable level, especially in platformer environments where collision boundaries, movement trajectories, and long-range reachability jointly determine feasibility.

To address this challenge, we propose a physics-aware PCGRL framework coordinating discrete level editing with continuous kinematic verification across three core components. First, we introduce a physics-aware state representation that embeds continuous kinematic constraints into the discrete MDP state space. Second, we design a curriculum-guided optimization strategy with informative feedback to improve exploration efficiency in long-horizon generation tasks. Finally, we employ a diversity-preserving synthesis mechanism to alleviate deterministic policy collapse after convergence.

The overall framework is illustrated in Figure~\ref{fig:framework}.

\subsubsection{Tri-Actor Execution Architecture}

To bridge discrete tile layout generation with continuous reachability constraints, our framework explicitly coordinates three distinct computational actors during each decision step:
\begin{enumerate}
    \item \textbf{Level-Editing Agent ($\pi_\theta$):} A parameterized policy operating over the discrete canvas representation, emitting localized cursor movements and tile modifications under a fixed editing budget.
    \item \textbf{Continuous Physics Simulator ($\mathcal{E}$):} A deterministic rigid-body simulation engine that models character kinematics, gravitational acceleration, and continuous bounding-box collision detection.
    \item \textbf{Multi-Template Kinematic Verifier ($\mathcal{V}$):} An A*-based trajectory planner that evaluates structural reachability on the updated canvas by exploring discrete jump-impulse templates under the motion rules of $\mathcal{E}$.
\end{enumerate}
During generation, the agent modifies the canvas, the verifier $\mathcal{V}$ assesses global reachability using the simulator $\mathcal{E}$, and the resulting kinematic trajectory is transformed into physical features that feed back into the agent's observation and reward.

\subsection{Physics-aware State Representation}

Existing PCGRL approaches commonly represent levels as discrete tile matrices, where each cell describes a local structural element. However, such representations ignore the underlying continuous dynamics of game characters. In physics-constrained platformers, a single tile transition may correspond to complex interactions between collision geometry, movement trajectories, and reachable regions.

To bridge this representation gap, we introduce a hybrid state representation that combines multi-channel spatial level features with explicit cursor localization:
\begin{equation}
s_t = (X_t, p_t^{\text{cur}})
\end{equation}
where $X_t \in \mathbb{R}^{H \times W \times d}$ denotes the $d$-channel spatial feature tensor of the current level, and $p_t^{\text{cur}} = (x_t^{\text{cur}}, y_t^{\text{cur}}) \in [0, W-1] \times [0, H-1]$ represents the discrete canvas coordinates of the editing cursor.

Specifically, the spatial tensor $X_t$ encodes heterogeneous level information across $d$ channels, including solid terrain occupancy, lethal spike layouts, terminal start/goal markers, and the physics-aware envelope channel $\Phi(\tau)$ derived from character motion constraints. Unlike purely symbolic tile matrices, this unified representation allows the policy network to directly perceive continuous kinematic reachability boundaries alongside local structural geometry during optimization.

\subsubsection{Action Space and Generation Budget}

We formulate level editing through a localized Turtle-style cursor positioned at $p_t^{\text{cur}}$, which sequentially modifies local regions of the canvas.

To maintain sample efficiency and avoid combinatorial explosion during policy exploration, the action space is formalized as a compact discrete set consisting of six actions:
\begin{equation}
\begin{aligned}
\mathcal{A} = \{ & \text{Left}, \text{Right}, \text{Up}, \text{Down}, \\
& \text{Toggle-Solid}, \text{Toggle-Spike} \}
\end{aligned}
\end{equation}
yielding an action space of cardinality $|\mathcal{A}| = 6$ (\texttt{Discrete(6)}). Directional actions shift the cursor by one grid unit along the corresponding axis, whereas \text{Toggle-Solid} and \text{Toggle-Spike} flip the occupancy of the targeted coordinate $p_t^{\text{cur}}$ between passable air and structural platform terrain or lethal spikes, respectively. Start and goal terminal markers are initialized by the environment and remain immutable throughout editing. Each episode terminates when cumulative steps reach the execution budget $T_{budget} = 1.5 \times (H \times W)$ or when full source-to-goal kinematic reachability is achieved.

\subsubsection{Character Kinematic Model and Abstraction Operator}

In physics-constrained platformers, discrete grid connectivity does not guarantee dynamic reachability. The character's continuous motion within simulator $\mathcal{E}$ is governed by discrete-time Euler integration with step size $\Delta t = 0.02\,\text{s}$:
\begin{align}
x(t + \Delta t) &= x(t) + v_x(t)\Delta t \\
y(t + \Delta t) &= y(t) + v_y(t)\Delta t \\
v_y(t + \Delta t) &= \max\left(v_y(t) - g\Delta t, -v_{term}\right)
\end{align}
where $g$ denotes gravitational acceleration (in $\text{tiles/s}^2$), $v_{term}$ represents the maximum terminal falling velocity bounding downward free-fall speed, and horizontal speed is restricted to $v_x(t) \in \{-v_{run}, 0, v_{run}\}$.A jump applies an instantaneous vertical impulse $v_y \leftarrow v_{jump}$. Mid-air steering reversals, double jumps, and hovering mechanics are strictly prohibited.

Given a continuous trajectory $\tau = \{(x(t), y(t))\}_{t=0}^T$ extracted by verifier $\mathcal{V}$, we define the continuous-to-discrete abstraction operator $\Phi: \tau \rightarrow \mathcal{S}_{phy}$ mapping continuous spatial occupancy onto the discrete canvas grid $\mathbb{R}^{H \times W}$:
\begin{equation}
\Phi(\tau)_{i,j} = \mathbb{I}\left( \text{Cell}(i, j) \cap \left[ \bigcup_{t=0}^T B(t) \right] \neq \emptyset \right)
\end{equation}
where $\text{Cell}(i, j) = [j, j+1] \times [i, i+1]$ denotes the coordinate boundary of tile $(i, j)$, $B(t)$ is the character bounding box at time $t$, and $\mathbb{I}(\cdot)$ is the indicator function. The generated occupancy grid $\Phi(\tau)$ is concatenated as an auxiliary feature channel into the state tensor $X_t$.

\subsubsection{Inflated Volume Envelope}

Directly modeling player motion as point-wise transitions is insufficient for
physics-constrained level generation, since collision events depend on the
entire geometric extent of the character. Therefore, we introduce the
Inflated Volume Envelope, which converts continuous movement trajectories into
discrete spatial occupancy regions by accumulating collision boundaries over
time.

For each intermediate frame along a movement primitive, the character
collision boundary is represented by an axis-aligned bounding region:

\begin{equation}
B_t=
[x_t^{min},x_t^{max}]
\times
[y_t^{min},y_t^{max}]
\end{equation}

where $x_t^{min}$ and $x_t^{max}$ denote the horizontal boundaries of the
character footprint, while $y_t^{min}$ and $y_t^{max}$ represent the vertical
boundaries at time step $t$.

The complete motion envelope is constructed by accumulating the discrete collision regions across the sampled trajectory frames, explicitly including the initial and terminal states:
\begin{equation}
E(\tau) = \left( \bigcup_{m=0}^{M} B(t_m) \right) \cup B(T), \quad t_m = m \cdot \Delta t_s
\end{equation}
where $B(t_m)$ denotes the character's axis-aligned bounding box at discrete timestamp $t_m$, and $M = \lfloor T / \Delta t_s \rfloor$ is the number of uniform sampling intervals.

To encourage sufficient temporal sampling density and substantially reduce the risk of missing intermediate collision regions (tunneling artifacts) along curved jump arcs, the sampling interval $\Delta t_s$ is bounded by:
\begin{equation}
\Delta t_s \le \frac{\min(w, h)}{2 \max\left(|v_{x,max}|, |v_{y,max}|\right)}
\end{equation}
where $w$ and $h$ denote the width and height of the character footprint, and $v_{x,max}, v_{y,max}$ denote the peak velocities under the kinematic model. This heuristic ensures at least $50\%$ axial overlap between consecutive sampled bounding boxes $B(t_m)$ and $B(t_{m+1})$ along the primary motion axes, mitigating the likelihood of skipping over narrow spikes or single-tile barriers without incurring the computational overhead of continuous Minkowski-sum swept hulls.

Furthermore, when the multi-template verifier $\mathcal{V}$ identifies multiple valid paths connecting the origin to the target, the trajectory $\tau^*$ exhibiting the minimal aggregate jump impulse is selected to construct $E(\tau^*)$, preventing artificial volumetric inflation of the envelope induced by redundant aerial looping or oscillatory detours.

\begin{figure}[t] 
  \centering
  \includegraphics[width=\linewidth]{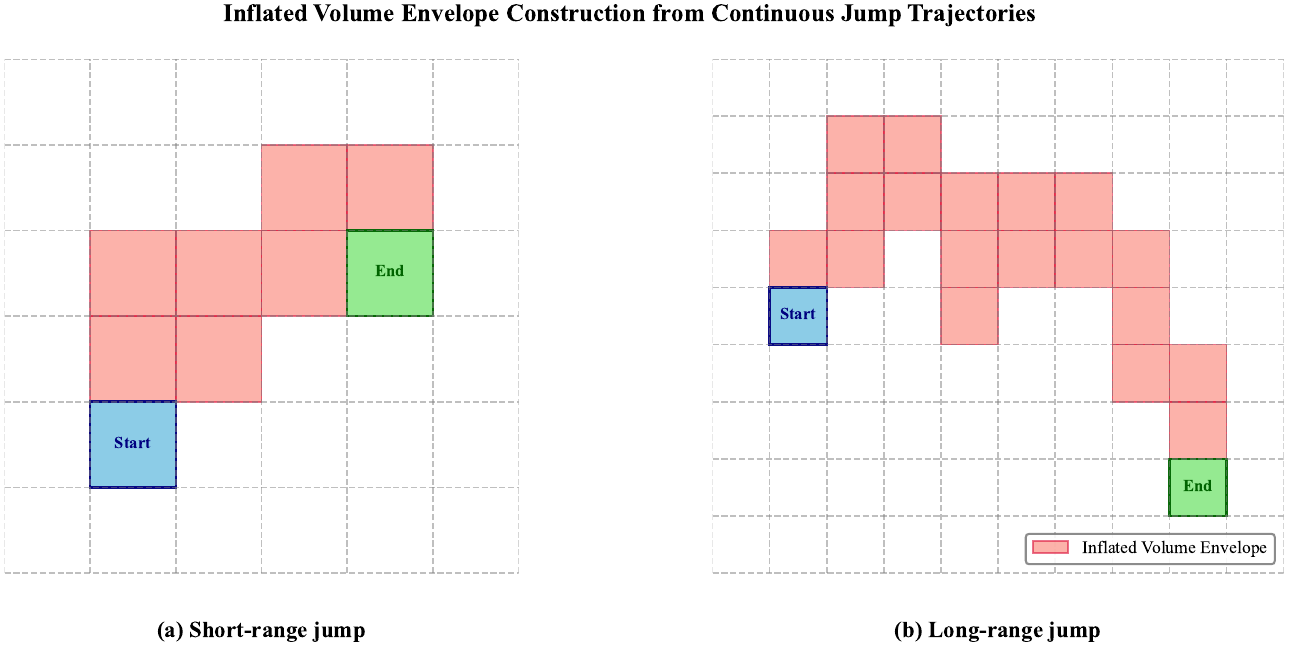} 
  \caption{
Illustration of the proposed Inflated Volume Envelope.
The continuous player motion trajectory is transformed into a discrete
occupancy region by accumulating collision boundaries over time.}
  \label{fig:jump_envelope_comparison}
\end{figure}

The overall construction process of the Inflated Volume Envelope is illustrated
in Figure~\ref{fig:jump_envelope_comparison}.

\begin{figure*}[t]
\centering
\includegraphics[width=\textwidth]{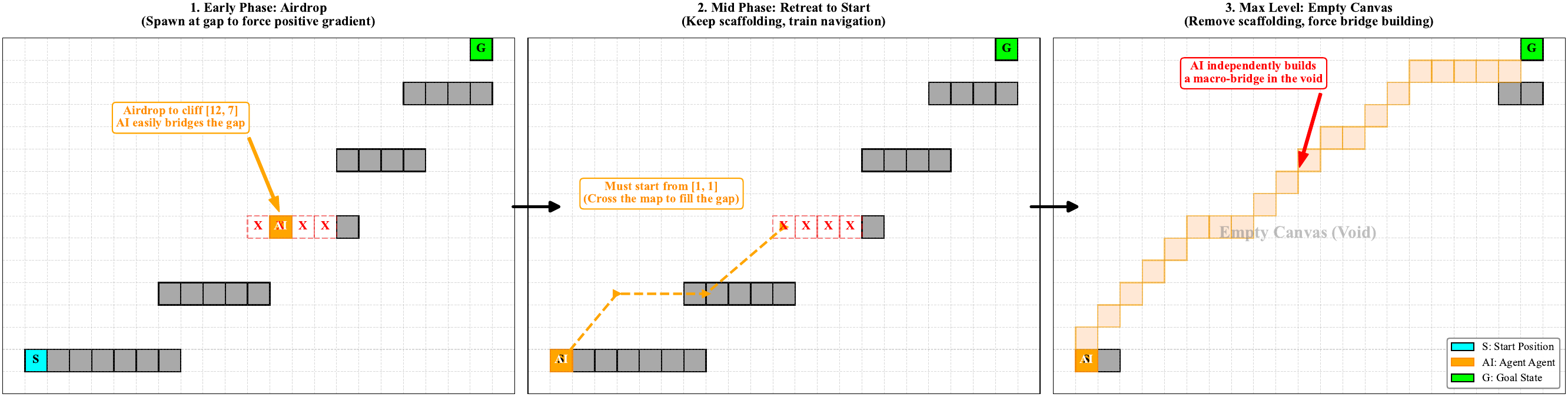}
\caption{
Illustration of the proposed forward extension curriculum.
The training process progressively transitions from scaffold-assisted
exploration (Early Phase), to navigation with partial structural guidance
(Mid Phase), and finally to autonomous level construction from an empty canvas
(Max Level).
}
\label{fig:curriculum}
\end{figure*}

\subsection{Curriculum-guided Optimization}

Although the proposed physics-aware representation improves the quality of
state estimation, directly optimizing large-scale level generation remains
challenging due to extremely sparse successful trajectories. In long-horizon
PCGRL tasks, meaningful rewards are often obtained only after completing an
entire playable path, making credit assignment difficult and causing policy
optimization to stagnate.

To address this issue, we introduce a curriculum-guided optimization strategy
that jointly improves learning signals through three complementary components:
(1) dense structural reward shaping, (2) intermediate state-transition
feedback, and (3) progressive forward curriculum expansion. These components
transform the original sparse generation objective into a structured
optimization process that gradually guides the agent from local feasibility
towards global level completion.

\subsubsection{Dense Reward and Structural Regularization}
Conventional PCGRL formulations relying on sparse terminal reachability rewards face severe gradient vanishing during early exploration. To provide continuous optimization guidance across both intermediate progress and global structural quality, we construct a multi-objective structured reward function:
\begin{equation}
R_t = \lambda_1 R_{play} + \lambda_2 R_{tight} + \lambda_3 R_{rhythm} + R_{flip} - \lambda_4 P_{garb}
\end{equation}
where $\lambda_1, \lambda_2, \lambda_3, \lambda_4 > 0$ denote non-negative weighting hyperparameters balancing terminal solvability, kinematic tightness, platforming cadence, and structural cleanliness, while $R_{flip}$ represents the intrinsic state-transition flip bonus formalized in Eq.~(\ref{eq:flip_reward}).

\paragraph{Solvability Reward ($R_{play}$).}
A binary terminal bonus awarded when the verifier establishes a complete feasible trajectory from the start to the target: $R_{play} = \mathbb{I}(\tau^* \text{ reaches goal})$.

\paragraph{Spatial Tightness Reward ($R_{tight}$).}
To prevent trivial solutions where the agent places platforms far apart in wide empty spaces, this term rewards levels requiring precise kinematic control near solid surfaces. Let $\mathcal{O}$ denote the solid terrain footprint, and let $\mathcal{N}(\mathcal{O}) = (\text{Dilate}(\mathcal{O}, \epsilon) \setminus \mathcal{O})$ define an exterior boundary margin of thickness $\epsilon = 1.0\,\text{tiles}$. The tightness metric measures the overlap between the safe envelope $E(\tau^*)$ and this interaction margin:
\begin{equation}
R_{tight} = \frac{|E(\tau^*) \cap \mathcal{N}(\mathcal{O})|}{|E(\tau^*)|}
\end{equation}
This formulation incentivizes layouts where the player must jump close to terrain edges without colliding with solid tiles.

\paragraph{Rhythmic Structural Regularization ($R_{rhythm}$).}
To encourage smooth platforming cadence rather than chaotic clusters of obstacles, $R_{rhythm}$ penalizes high spatial variance in the distance between consecutive jump launch coordinates $\{d_k\}$:
\begin{equation}
R_{rhythm} = \exp\left( - \frac{\text{Var}(\{d_k\})}{\sigma_d^2} \right)
\end{equation}
Policy entropy regularization is removed from the extrinsic reward formulation and incorporated directly into the PPO actor optimization objective.

\paragraph{Normalized Structural Penalty ($P_{garb}$).}
To discourage the generation of disconnected floating clutter and maintain invariant penalty scaling across varying canvas resolutions (from $15 \times 12$ to $25 \times 16$), functionally redundant structures are penalized with canvas-area normalization:
\begin{equation}
P_{garb} = \frac{N_{invalid}}{H \times W}
\end{equation}
where $N_{invalid}$ denotes the count of isolated or floating solid tiles that neither provide physical foothold support along the reachable trajectory margin $\mathcal{N}(E(\tau_t^{\text{reach}}))$ nor belong to any connected structural terrain component supporting traversable paths.

\subsubsection{Forward Extension Curriculum}

Although dense feedback improves local optimization, directly learning complete
large-scale level generation remains challenging. In physics-constrained
platformers, a successful level requires a long sequence of coordinated
decisions, where early actions may influence the feasibility of future
structures. Therefore, a randomly initialized agent faces an extremely sparse
search space, making it difficult to discover meaningful generation behaviors
through end-to-end optimization.

To address this challenge, we introduce a forward extension curriculum that
gradually expands the generation objective from locally achievable structures
to complete level synthesis. Instead of requiring the agent to solve the full
generation task from the beginning, the curriculum progressively removes
external scaffolding and increases the spatial dependency of the task.

As illustrated in Figure~\ref{fig:curriculum}, the proposed curriculum
consists of three progressive stages. In the early phase, the agent is placed
near an intermediate reachable region, allowing it to quickly experience
successful transitions and establish initial optimization signals. In the
mid phase, the agent is returned to the original starting position while
partial structural scaffolding is preserved. This stage encourages the policy
to learn navigation across longer distances while maintaining informative
feedback. Finally, in the maximum-level phase, all auxiliary structures are
removed and the agent is required to autonomously construct complete level
topologies from an empty canvas.

This progressive expansion transforms the original sparse generation problem
into a sequence of increasingly difficult sub-problems, enabling the agent to
first acquire physically feasible behaviors and subsequently optimize global
level structures and diversity.

To remove heuristic ambiguity and ensure reproducible progression, transitions between the three curriculum stages are strictly governed by quantitative performance gating:
\begin{equation}
\text{Stage } k \rightarrow k+1 \iff \bar{S}_{win} \ge \delta_k \quad \land \quad F_t \ge \rho_k W
\end{equation}
where $\bar{S}_{win} = \frac{1}{N_{win}} \sum_{m=t-N_{win}+1}^t \mathbb{I}(\tau_m^* \text{ is solvable})$ denotes the rolling solvability success rate evaluated over a fixed window of $N_{win} = 100$ preceding training episodes, $F_t$ represents the exploration frontier memory distance, $W$ is the canvas width, and $\delta_k, \rho_k \in (0, 1)$ denote the stage progression thresholds. 

Specifically, the environment transitions from the Early Phase to the Mid Phase when the agent demonstrates local physical stability ($\delta_1 = 0.80$, $\rho_1 = 0.40$), retracting half of the localized scaffolding. The transition from the Mid Phase to the Max Level phase occurs once the agent sustains long-horizon reachability across the canvas ($\delta_2 = 0.85$, $\rho_2 = 0.90$), at which point all external structural scaffolds are completely withdrawn. If the rolling solvability drops significantly below acceptable convergence stability ($\bar{S}_{win} < \delta_k - 0.20$), the curriculum controller retains the current stage without regression, stabilizing credit assignment across expanding action horizons.

\subsubsection{Exploration Frontier Memory}

During curriculum optimization, exploration progress may fluctuate due to the
stochastic nature of reinforcement learning. To prevent previously discovered
feasible regions from being repeatedly lost, we maintain a historical
exploration frontier that records the maximum validated progression achieved
during training:

\begin{equation}
F_t=\max(F_{t-1},d_t)
\end{equation}

where $d_t$ represents the currently achieved reachable distance.

The monotonic frontier provides a stable reference for curriculum expansion.
Instead of relying only on instantaneous rewards, the agent preserves previous
exploration achievements and gradually extends the reachable region toward the
final objective.

\subsubsection{State-Transition Flip Feedback}

Long-horizon PCGRL tasks suffer from severe credit assignment degradation because terminal playability verification provides meaningful reward signals only after the entire canvas layout is successfully synthesized. To supply informative intermediate feedback along extended editing sequences, we partition the canvas horizontally into $K$ uniform longitudinal segments $\{C_1, C_2, \dots, C_K\}$.

For each intermediate segment $i \in \{1, \dots, K\}$, we define a binary reachability state variable $v_{i,t} \in \{0, 1\}$ at editing step $t$:
\begin{equation}
v_{i,t} = \mathbb{I}\!\left( \exists (x, y) \in \tau_t^{\text{reach}} \;\text{s.t.}\; x \in \left[\tfrac{(i-1)W}{K}, \tfrac{iW}{K}\right) \right)
\end{equation}
where $\tau_t^{\text{reach}}$ denotes the maximum-progress reachable trajectory extracted from the explored search tree of verifier $\mathcal{V}$ at step $t$ (representing the furthest reachable frontier trajectory prior to full level completion, and coinciding with the complete optimal trajectory $\tau_t^*$ once the goal is connected), and $\mathbb{I}(\cdot)$ is the indicator function.

To capture structural progress dynamically, a state-transition flip event $\Delta v_{i,t}$ is emitted exclusively at the step where an unreachable segment transitions into a reachable state:
\begin{equation}
\Delta v_{i,t} = \max\left(0, v_{i,t} - v_{i,t-1}\right)
\end{equation}

When a forward reachability flip is detected ($\Delta v_{i,t} = 1$), an immediate intrinsic progress bonus is injected into the step-wise reward:
\begin{equation}
\label{eq:flip_reward}
R_{flip}(s_t, a_t) = \sum_{i=1}^K \eta_i \Delta v_{i,t}
\end{equation}
where $\eta_i > 0$ denotes a monotonically increasing scale weight assigning higher credit to deeper spatial milestones.

Unlike static proximity heuristics or ungrounded dense shaping, this event-driven flip feedback guarantees that credit is assigned only when an editing action directly establishes new continuous kinematic connectivity across intermediate topological boundaries.

\subsubsection{Progressive Stage Expansion}

The curriculum controller gradually increases task complexity according to the
agent's exploration capability. Early stages emphasize local physical
feasibility, while later stages introduce larger spatial dependencies and
global topology requirements.

This design follows a coarse-to-fine optimization principle: the agent first
learns reliable movement and construction patterns under constrained
conditions, and then transfers these learned behaviors to unconstrained
large-scale generation.

By progressively expanding the optimization horizon, the proposed curriculum
reduces the difficulty of discovering long-range valid trajectories and
improves scalability for complex physics-constrained environments.

\subsubsection{Scaling Analysis of Long-Horizon Exploration}

The fundamental difficulty of large-scale physics-constrained generation lies
in the interaction between increasing decision horizons and sparse feasible
solutions. As the canvas size grows, the agent must explore longer sequences
of editing actions while simultaneously satisfying spatial connectivity,
collision constraints, and dynamic reachability.

In conventional end-to-end PCGRL frameworks, these requirements are coupled
into a single optimization objective. Consequently, the probability of
discovering successful trajectories during early exploration decreases as the
generation space expands, resulting in unstable learning behavior.

The proposed optimization strategy alleviates this difficulty by introducing
progressive objectives and intermediate feedback. The forward curriculum
reduces the initial search difficulty by constraining early learning stages,
while state-transition feedback provides additional optimization signals along
the generation trajectory.

Empirical results in experiments further validate that the
proposed framework maintains stable convergence across different canvas
scales, while direct end-to-end optimization suffers from degraded
solvability as the generation horizon increases.

\section{Experiments}

\begin{table}[t]
\centering
\small
\caption{System Hyperparameters, Kinematic Constants, and Training Configuration.}
\label{tab:system_hyperparameters}
\resizebox{\columnwidth}{!}{%
\begin{tabular}{ll}
\toprule
\textbf{Parameter / Notation} & \textbf{Value / Setting} \\
\midrule
\multicolumn{2}{l}{\textit{\textbf{Avatar Kinematics \& Simulator}}} \\
Gravitational acceleration ($g$) & $9.8\,\text{tiles/s}^2$ \\
Running horizontal speed ($v_{run}$) & $2.0\,\text{tiles/s}$ \\
Jump vertical impulse ($v_{jump}$) & $7.5\,\text{tiles/s}$ \\
Terminal falling velocity ($v_{term}$) & $12.0\,\text{tiles/s}$ \\
Avatar collision footprint ($w \times h$) & $0.484 \times 0.66\,\text{tiles}$ \\
Simulation integration step ($\Delta t$) & $0.02\,\text{s}$ ($50\,\text{Hz}$) \\
\midrule
\multicolumn{2}{l}{\textit{\textbf{PPO Algorithm \& Network}}} \\
Policy architecture & \texttt{MultiInputPolicy} \\
Optimization learning rate ($\alpha$) & $3 \times 10^{-4}$ \\
Discount factor ($\gamma$) & $0.99$ \\
GAE parameter ($\lambda_{gae}$) & $0.95$ \\
Policy clipping threshold ($\epsilon$) & $0.20$ \\
Policy entropy coefficient ($\beta$) & $0.03$ \\
Rollout buffer / Mini-batch & $2048$ / $64$ \\
Action space & $6$-action Discrete (\texttt{Discrete(6)}) \\
Observation space & \begin{tabular}[t]{@{}l@{}}Dict observation: multi-channel grid \\ (incl. physics-aware envelope) $+$ \\ agent cursor coordinates $p_t^{\text{cur}}$\end{tabular} \\
\bottomrule
\end{tabular}%
}
\end{table}

\subsection{Experimental Setup}

\paragraph{Environment.}

We evaluate our framework on a physics-constrained 2D platformer generation environment, where an agent constructs playable levels through sequential editing actions. Each generated level is evaluated by an embedded physics simulator and a multi-template A* verifier to determine physical reachability.

The environment contains rigid-body collision constraints, nonlinear jumping dynamics, and long-range spatial dependencies, making it a challenging benchmark for PCGRL systems.

Unlike conventional tile-based generation tasks, successful generation requires not only structural connectivity but also physically feasible movement trajectories.

\begin{figure}[t]
  \centering
  \begin{subfigure}[b]{0.48\linewidth}
    \centering
    \includegraphics[width=\linewidth]{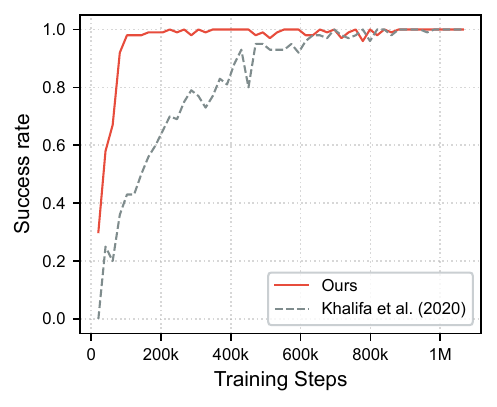}
    \caption{$15 \times 12$}
    \label{fig:map_1512}
  \end{subfigure}
  \hfill
  \begin{subfigure}[b]{0.48\linewidth}
    \centering
    \includegraphics[width=\linewidth]{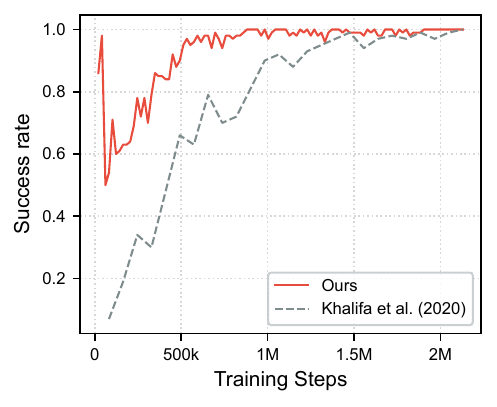}
    \caption{$18 \times 14$}
    \label{fig:map_1814}
  \end{subfigure}

  \vspace{1.5mm} 

  \begin{subfigure}[b]{0.48\linewidth}
    \centering
    \includegraphics[width=\linewidth]{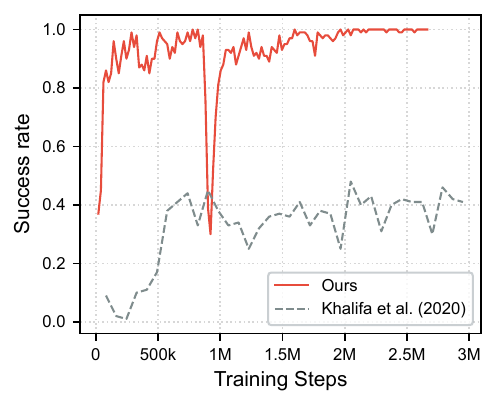}
    \caption{$21 \times 15$}
    \label{fig:map_2115}
  \end{subfigure}
  \hfill
  \begin{subfigure}[b]{0.48\linewidth}
    \centering
    \includegraphics[width=\linewidth]{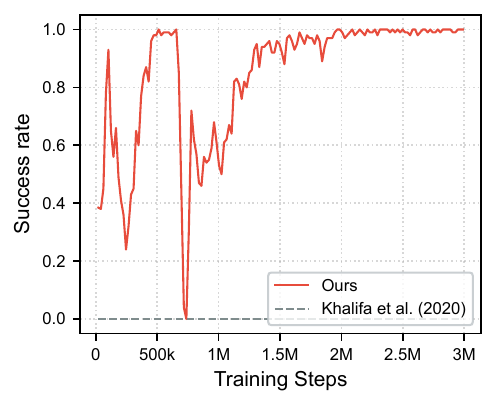}
    \caption{$25 \times 16$}
    \label{fig:map_2516}
  \end{subfigure}

  \caption{Comparison of training success rates across four different canvas dimensions: (a) $15 \times 12$, (b) $18 \times 14$, (c) $21 \times 15$, and (d) $25 \times 16$.}
  \label{fig:multi_scale_comparison}
\end{figure}

\paragraph{Baselines.}

We compare our method against the representative PCGRL framework proposed by Khalifa et al. (2020), which formulates level generation as a reinforcement learning problem through direct end-to-end optimization.

This baseline represents the standard PCGRL paradigm without physics-aware state abstraction or curriculum-guided exploration.

\paragraph{Implementation Details.}
The complete kinematic parameters governing the physics simulator and the training hyperparameters for policy optimization are summarized in Table~\ref{tab:system_hyperparameters}.The character locomotion is integrated at a simulation step of $\Delta t = 0.02\,\text{s}$ ($50\,\text{Hz}$) under gravitational acceleration $g = 9.8\,\text{tiles/s}^2$, a terminal falling velocity of $v_{term} = 12.0\,\text{tiles/s}$, and a bounding collision footprint of $0.484 \times 0.66\,\text{tiles}$. For policy optimization, we employ PPO with a \texttt{MultiInputPolicy} architecture to process the heterogeneous dictionary observation space comprising the multi-channel grid representation (incorporating the physics-aware envelope channel) and the agent cursor coordinates $p_t^{\text{cur}}$. All agents are trained using the Adam optimizer with a learning rate of $\alpha = 3 \times 10^{-4}$, a discount factor of $\gamma = 0.99$, and an entropy regularization coefficient of $\beta = 0.03$ over rollout buffers of 2048 steps.

\paragraph{Evaluation Metrics.}

We evaluate generated levels from three perspectives:

\textbf{Level Solvability.}
The success rate measures the proportion of generated levels that can be completed by the embedded physics-based verification agent.

\textbf{Optimization Stability.}
Training convergence is evaluated by monitoring success rate evolution during reinforcement learning.

\textbf{Structural Diversity.}
Generated contents are analyzed through topology variation, kinematic clearance margins, and trajectory diversity metrics.

All experiments are conducted across multiple canvas resolutions to evaluate scalability under increasing generation complexity.

\begin{figure}[t]
  \centering
  \includegraphics[width=\linewidth]{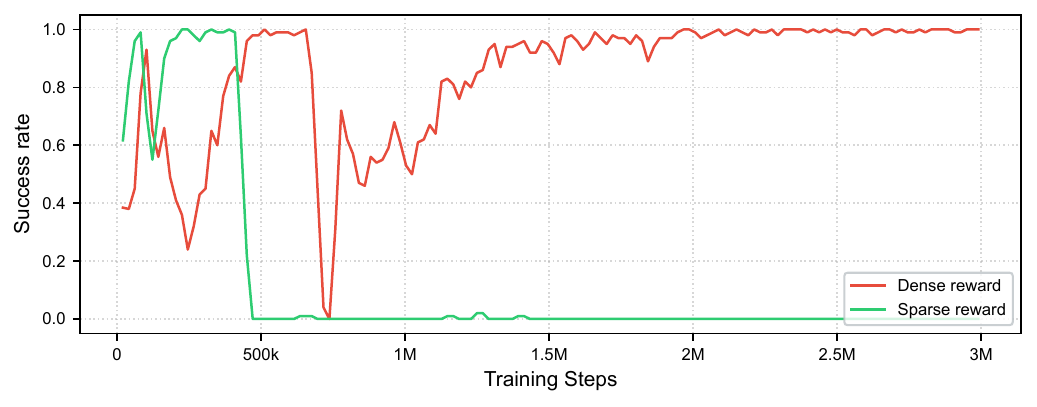}
  \caption{Training curves comparing dense reward shaping against sparse reward baseline over 3M training steps.}
  \label{fig:reward_ablation}
\end{figure}

\begin{figure*}[t]
  \centering
  \begin{subfigure}[c]{0.21\textwidth}
    \centering
    \includegraphics[width=\linewidth]{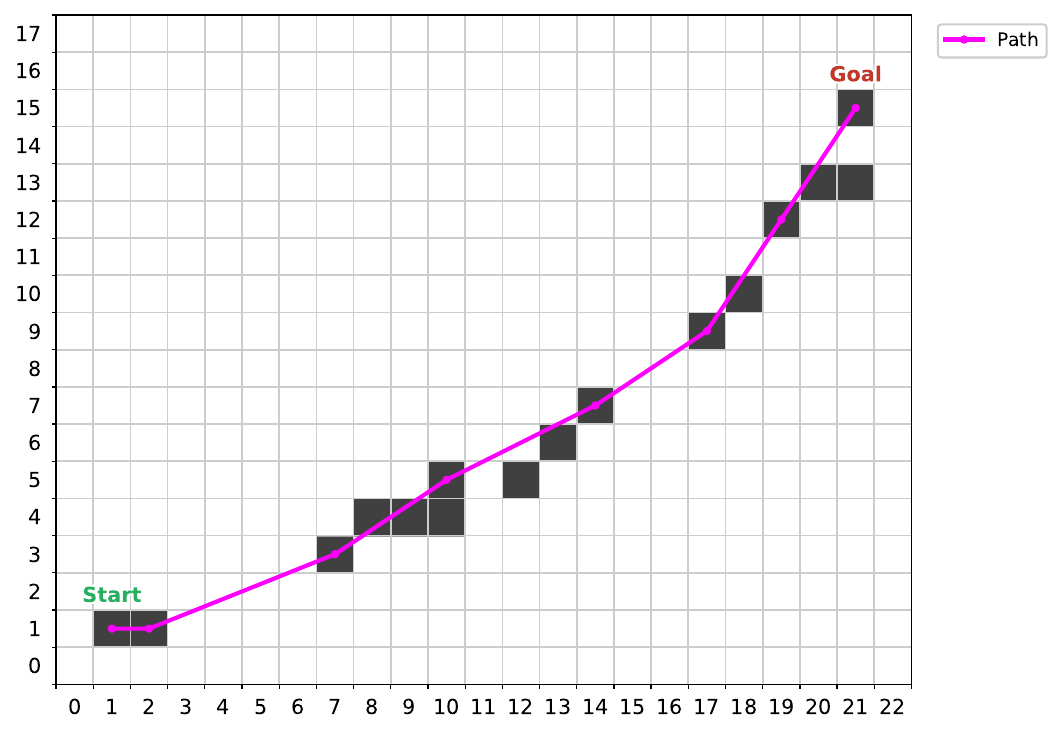}
    \caption{Deterministic}
    \label{fig:synth_a}
  \end{subfigure}%
  \hfill
  \begin{subfigure}[c]{0.03\textwidth}
    \centering
    \Large $\rightarrow$
  \end{subfigure}%
  \hfill
  \begin{subfigure}[c]{0.21\textwidth}
    \centering
    \includegraphics[width=\linewidth]{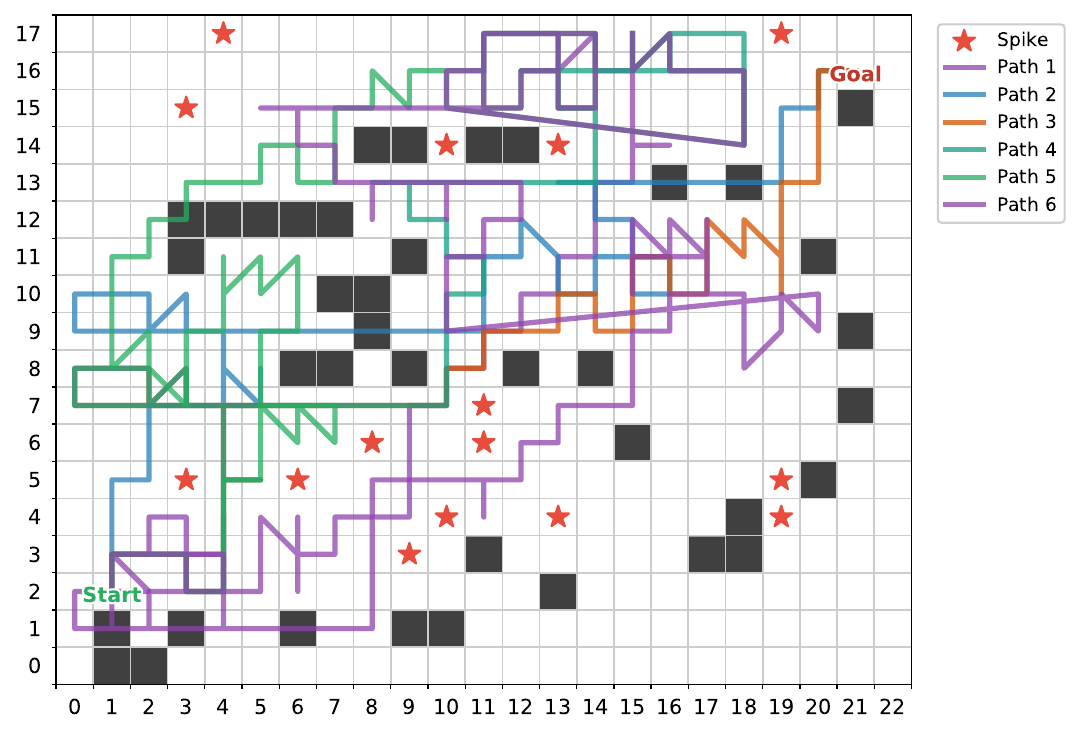}
    \caption{Stochastic \#1}
    \label{fig:synth_b}
  \end{subfigure}%
  \hfill
  \begin{subfigure}[c]{0.21\textwidth}
    \centering
    \includegraphics[width=\linewidth]{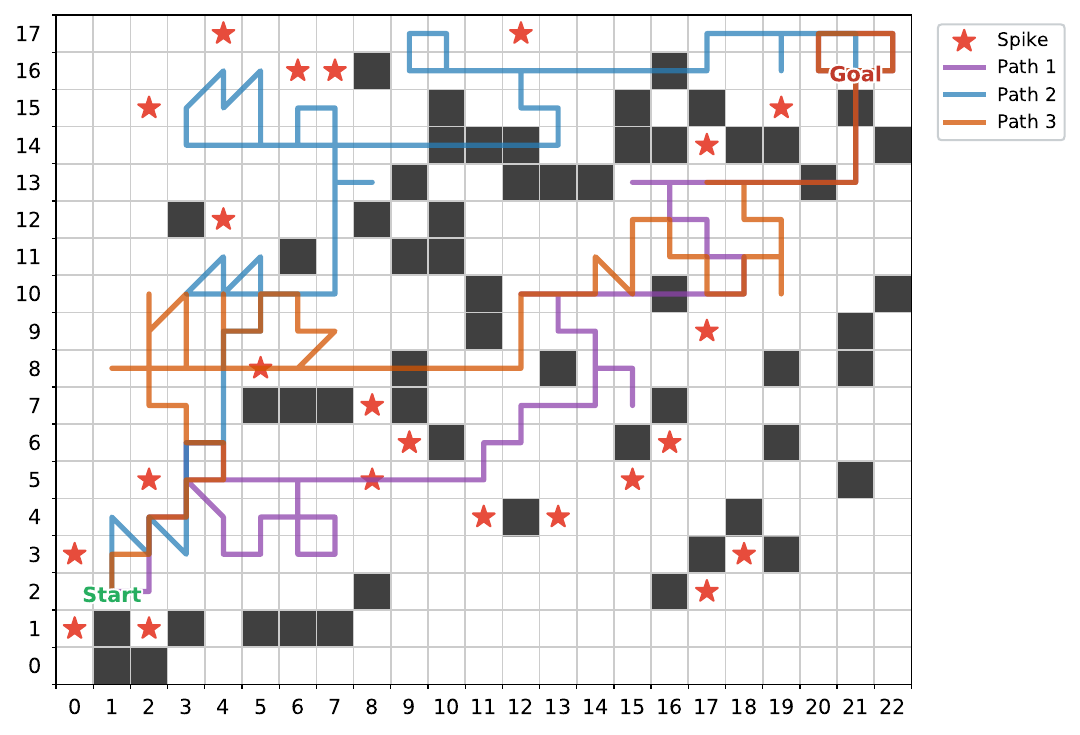}
    \caption{Stochastic \#2}
    \label{fig:synth_c}
  \end{subfigure}%
  \hfill
  \begin{subfigure}[c]{0.21\textwidth}
    \centering
    \includegraphics[width=\linewidth]{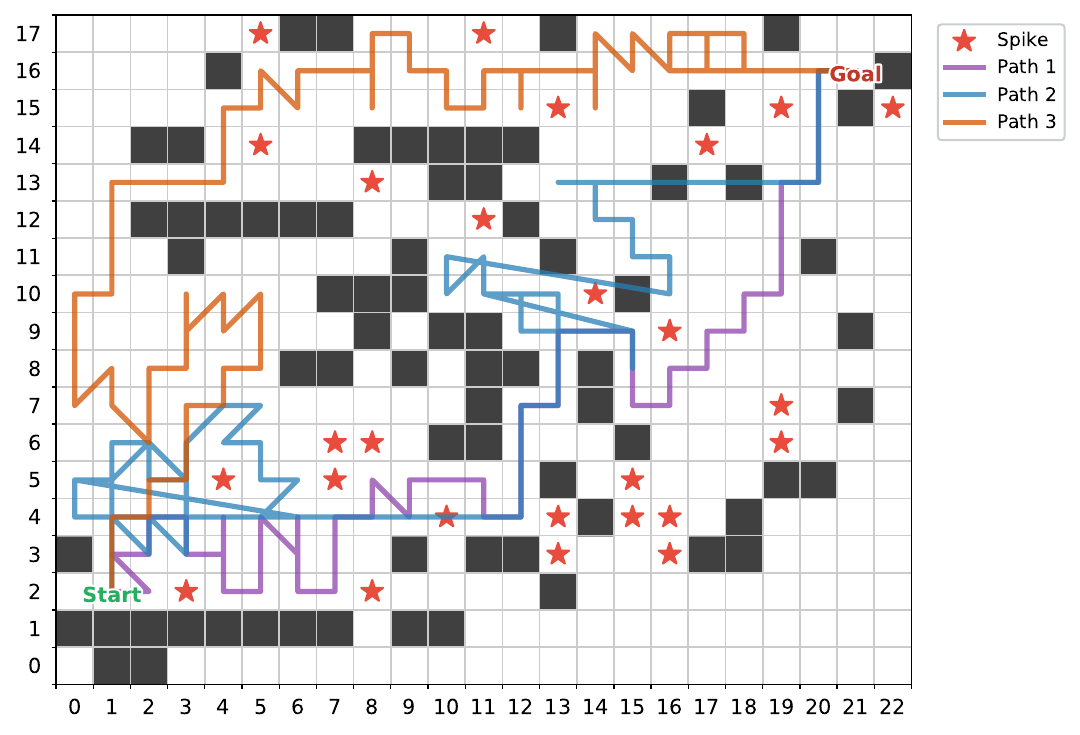}
    \caption{Stochastic \#3}
    \label{fig:synth_d}
  \end{subfigure}

  \caption{Comparison of synthesis modes: (a) deterministic inference yielding a single monotone path, and (b)--(d) diverse level layouts with multiple feasible routes synthesized via high-entropy stochastic inference.}
  \label{fig:stochastic_synthesis_comparison}
\end{figure*}

\subsection{Comparison with Existing PCGRL}
\label{sec:comparison}

We compare our framework with the conventional PCGRL approach to evaluate
whether the proposed physics-aware representation and curriculum-guided
optimization improve large-scale physics-constrained level generation.

Figure~\ref{fig:multi_scale_comparison} summarizes the success rate evolution
of our method and the baseline under progressively increasing canvas scales.

On relatively small environments (15$\times$12), both approaches are able to
discover feasible layouts through reinforcement learning optimization.
However, our method achieves faster convergence and reaches higher solvability
earlier, benefiting from the additional guidance provided by the
physics-aware state representation and intermediate structural feedback.

As the generation scale increases, the performance gap becomes increasingly
significant. The baseline PCGRL framework exhibits degraded optimization
performance on larger canvases, where the agent must discover long sequences
of physically valid editing operations with limited intermediate feedback.
This results in unstable exploration and lower final success rates.

In contrast, our framework maintains reliable optimization across different
canvas scales. Although temporary performance fluctuations appear in larger
environments, these changes correspond to curriculum transitions where the
generation objective progressively expands from constrained intermediate
structures to complete large-scale layouts. After adapting to the expanded
search space, the policy consistently recovers and achieves high success
rates.

The results demonstrate that the proposed framework effectively improves the
scalability of PCGRL for large-scale physics-constrained level generation by
combining physics-aware state representation with curriculum-guided
exploration.

\subsubsection{Effect of Curriculum Feedback}

We first investigate the contribution of intermediate state-transition feedback.

The comparison removes the proposed dense feedback mechanism while keeping the same curriculum schedule.

As shown in Figure~\ref{fig:reward_ablation}, removing intermediate feedback leads to unstable optimization.

Although the sparse-reward variant achieves high performance in early stages due to simplified initial conditions, it fails when the curriculum progresses toward long-horizon generation.

Without local transition signals, the agent receives meaningful feedback only after complete level completion, resulting in inefficient credit assignment.

In contrast, our full model maintains stable convergence by transforming global completion objectives into incremental structural improvements.

\subsubsection{Effect of Diversity Preservation}

Finally, we investigate the impact of the proposed diversity-preserving
synthesis strategy.

Although reinforcement learning policies can achieve high playability after
convergence, deterministic inference often favors a limited set of
high-reward solutions, resulting in repetitive topology patterns and reduced
structural diversity.

Figure~\ref{fig:stochastic_synthesis_comparison} compares a fully converged deterministic
policy with several levels generated using stochastic synthesis. The
deterministic policy produces a highly consistent but monotonous layout,
whereas stochastic inference generates diverse path structures and spatial
organizations while maintaining physical feasibility.

These results demonstrate that the proposed synthesis strategy effectively
mitigates policy-induced mode collapse and provides a better balance between
playability and structural diversity.

\subsection{Qualitative and Diversity Analysis}

Beyond quantitative performance metrics, we further analyze the qualitative
characteristics of generated levels.

Figure~\ref{fig:stochastic_synthesis_comparison} presents representative levels generated by
our framework. The results demonstrate that the proposed method can produce
varied spatial organizations, including different platform configurations,
path topologies, and difficulty distributions, while preserving physical
playability.

To further evaluate the generated content, we analyze the relationship
between structural diversity, action entropy, and path trajectory dispersion. The results show
that excessive deterministic convergence reduces topology variation, whereas
controlled stochastic generation achieves a more favorable balance between
diversity and solvability.

\section{Conclusion}

In this work, we investigated the structural bottlenecks of Procedural Content Generation via Reinforcement Learning (PCGRL) in physics-constrained 2D platformers. To bridge the gap between discrete grid-based editing actions and continuous kinematic reachability, we presented a physics-aware generation framework combining the Inflated Volume Envelope with a forward extension curriculum. By projecting the avatar's swept collision footprint into the discrete MDP state and guiding agent exploration from localized scaffolded sub-goals to autonomous layout synthesis, our framework significantly mitigates sparse-reward collapse and credit assignment failure across extended canvas dimensions. Furthermore, pairing high-entropy stochastic inference with parallel rejection filtering alleviates the behavioral stagnation characteristic of deterministic policy convergence, yielding diverse and physically traversable level configurations.

\subsection{Limitations}

Despite the empirical gains across multi-scale benchmarks, several foundational limitations of this framework warrant critical discussion:

\subsubsection{Content Alphabet and Scale Boundary.}
The current experimental validation is confined to canvas resolutions up to $25 \times 16$ and a discrete tile set consisting of solid terrain, lethal spikes, and terminal markers. While this configuration serves as a rigorous, controlled testbed for isolating rigid-body kinematic bottlenecks, it does not incorporate dynamic hazards (e.g., moving enemies, falling platforms) or multi-room global transitions. Extending the action space to richer visual semantics without inducing combinatorial explosion remains an open challenge.

\subsubsection{Degenerate Heuristics and Trivial Paths.}
As observed during deterministic convergence, dense proximity and playability rewards can inadvertently incentivize monotonic staircase structures---the path of least resistance under fixed jump impulses. Although stochastic sampling disrupts this local optimum, the post-hoc rejection filter introduces computational overhead during inference, and acceptance rates depend heavily on temperature calibration and sample candidate budgets.

\subsubsection{Kinematic Simplifications.}
The Inflated Volume Envelope relies on discrete temporal sampling of axis-aligned bounding boxes (AABB) under fixed horizontal velocities and deterministic jump arcs. Environments featuring variable impulse jumps, mid-air steering modulation, or rotating collision bodies would require continuous swept-volume hulls or dynamic Minkowski sums to prevent tunneling artifacts between sampled frames.

\subsection{Future Work}

These constraints highlight promising avenues for subsequent research.First, inspired by latent illumination frameworks \citep{fontaine2021illuminating, bhatt2022deep}, we plan to embed Quality-Diversity (QD) objectives directly into the reinforcement learning training loop rather than relying exclusively on post-hoc rejection filtering, thereby explicitly optimizing for behavioral and topological divergence.

Second, integrating macro-level topological priors (e.g., via graph neural networks or pretrained spatial models) with low-level kinematic envelope verifiers offers a viable pathway toward generating arbitrarily long, non-linear multi-path adventures. Finally, scaling the swept collision envelope formulation to 3D voxel-based obstacle courses represents a natural progression for continuous-physics PCG systems.

\small

\bibliography{aaai2027} 

\end{document}